# Event-based Vision meets Deep Learning on Steering Prediction for Self-driving Cars

Ana I. Maqueda[1], Antonio Loquercio[2], Guillermo Gallego[2],
Narciso García[1], and Davide Scaramuzza[2].
[1]Grupo de Tratamiento de Imágenes, Universidad Politécnica de Madrid, Spain
[2]Dept. of Informatics and Neuroinformatics, University of Zurich and ETH Zurich, Switzerland

**Abstract**

*Event cameras are bio-inspired vision sensors that naturally capture the dynamics of a scene, filtering out redundant information. This paper presents a deep neural network approach that unlocks the potential of event cameras on a challenging motion-estimation task: prediction of a vehicle's steering angle. To make the best out of this sensor–algorithm combination, we adapt state-of-the-art convolutional architectures to the output of event sensors and extensively evaluate the performance of our approach on a publicly available large scale event-camera dataset (≈1000 km). We present qualitative and quantitative explanations of why event cameras allow robust steering prediction even in cases where traditional cameras fail, e.g. challenging illumination conditions and fast motion. Finally, we demonstrate the advantages of leveraging transfer learning from traditional to event-based vision, and show that our approach outperforms state-of-the-art algorithms based on standard cameras.*

## Multimedia Material

A video accompanying this paper can be found at:
https://youtu.be/_r_bsjkJTHA

## 1. Introduction

Event cameras, such as the Dynamic Vision Sensor (DVS) [1], are bio-inspired sensors that, in contrast to traditional cameras, do not acquire full images at a fixed framerate but rather have independent pixels that output only *intensity changes* (called "events") asynchronously at the time they occur. Hence, the output of an event camera is not a sequence of images but a stream of asynchronous events. Event cameras have multiple advantages over traditional cameras: very high temporal resolution (microseconds), very high dynamic range (HDR) (140 dB) and low power and bandwidth requirements. Moreover, since events are generated by moving edges in the scene, event cameras are natural *motion detectors* and automatically filter out any temporally-redundant information. Due to their principle of operation and unconventional output, event cameras represent a paradigm shift in computer vision, and so, new algorithms are needed to exploit their capabilities. Indeed, event cameras present many advantages in all tasks related to motion estimation [2, 3, 4, 5, 6].

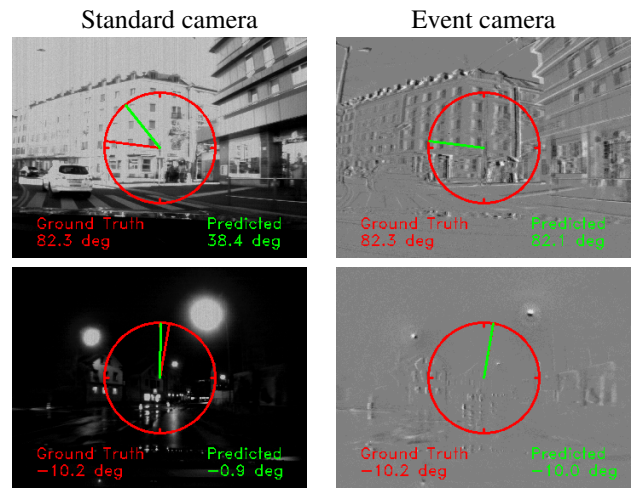

Figure 1: Steering angle regression performance on grayscale frames (first column) and on event camera data (second column). The first row shows a sunny day where visual features can be extracted from grayscale frames. However the camera saturation and the lack of temporal information makes the network predict a wrong steering angle. The second row shows a night scene, from which the network hardly predicts the correct steering angle when using grayscale frames. Our method accurately predicts a steering angle by combining event data and deep learning in both scenarios.

Recently, deep learning (DL) algorithms were shown to perform well on many applications in the field of motion estimation [7, 8, 9]. In this work, we propose to unlock the potential of event cameras through a DL-based solution, and showcase the power of this combination on the challenging task of steering angle prediction. As already pointed out by previous work [9], learning-based approaches will be ultimately needed to handle complex scenarios and, more importantly, corner cases in which self-driving cars will maneuver. However, the goal of this work is not to develop a framework to actually control an autonomous car or robot, as already proposed in [10]. On the contrary, we aim at understanding how learning-based approaches to motion-estimation tasks could benefit from the natural response of event cameras to motion, their inherent data redundancy reduction, high speed and very high dynamic range. We show that the ability of event cameras to capture the dynamics of a scene at low-latency combined with specifically-designed neural networks outperforms state-of-the-art systems which are based on standard cameras (Fig. 1).

Overall, this paper makes the following contributions:

- We show the first large-scale ($\approx$ 1 million images covering over 1000 km) application of deep learning to event-based vision on a regression task. Additionally, we provide results and explanations on why an event camera is better suited to motion-estimation tasks than a traditional camera.
- We adapt state-of-the-art convolutional architectures [11] to the output of event cameras. Furthermore, we show that it is possible to leverage transfer learning from pre-trained convolutional networks on classification tasks [12], even if the networks were trained on frames collected by traditional cameras.
- We prove the validity of our methodology through an extensive set of qualitative and quantitative experiments outperforming state-of-the-art systems on a publicly available dataset.

The rest of the paper is organized as follows. Section 2 reviews related work on the problem. Section 3 describes the proposed methodology, whose performance is extensively evaluated in Sections 4 and 5. Results are discussed in Section 6, and conclusions are drawn in Section 7.

## 2. Related Work

Developing robust policies for autonomous driving is a challenging research problem. Highly engineered, modular systems demonstrated incredible performance in both urban and off-road scenarios [13]. Another approach to the problem is to directly map visual observations to control actions, tightly coupling the perception and control parts of the problem. The first attempt to learn a visuomotor policy was done with ALVINN [14], where a shallow network was used to predict actions directly from images. Even though it only succeeded in simple scenarios, it suggested the potential of neural networks for autonomous navigation. More recently, NVIDIA used a CNN to learn a driving policy from video frames [15]. In spite of being a very simple approach, the learned controls were able to drive a car in basic scenarios. Afterwards, several research efforts have been spent to learn more robust perception-action models [9, 8] to cope with the diversity of visual appearance and unpredictability usually encountered in urban environments. Xu et al. [9] proposed to leverage large-scale driving video datasets and to do transfer learning to generate more robust policies. The model showed good performance but was limited to only a set of discrete actions and was susceptible to failures in undemonstrated regions of the policy space. In [8] the authors proposed a method to directly regress steering angles from frames while providing an interpretable policy. However, regarding performance, very little improvement was achieved with respect to [15].

All previous methods operate on images acquired by traditional frame-based cameras. In contrast, we propose to learn policies based on the data produced by event cameras (asynchronous, pixel-wise brightness changes with very low latency and high dynamic range), which naturally respond to *motion* in the scene.

The capabilities of event cameras to provide rich data for solving pattern recognition problems has been initially shown in [16, 17, 18, 19, 10]. In all these problems, machine learning algorithms were applied on data acquired by an event camera to solve classification problems, and were generally trained and tested on datasets of limited size. For example [16, 17, 18] use neural networks on event data to recognize cards of a deck (4 classes), faces (7 classes) or characters (36 classes). A similar case is that of [19], where a network is trained to recognize three types of gestures (rocks, papers, scissors) in dynamic scenes. So far, estimation problems in which the unknown variable is continuous were tackled by discretization, i.e., the solution space was partitioned into a finite number of classes. This is the case, for example, of the predator-prey robots in [10], where a network trained on the combined input of events and grayscale frames from a Dynamic and Active-pixel Vision Sensor (DAVIS) [20] produced one of four outputs: the prey is on the left, center, or right of the predator's field of view (FOV), or it is not visible in the FOV. Another example is that of the optical flow estimation method in [21], where the network produced motion vectors from a set with 8 different directions and 8 different speeds (i.e., 64 classes).

As opposed to the classification approximation of *all* previous methods, this paper addresses a continuous estimation problem (steering angle prediction) from a *regression* point of view. Hence, we are the first to tackle continuous estimation problems with event cameras in a principled way without resorting to partitioning the solution space; the angles produced by our network can take any value, not just discrete ones, in the range $[-180°, 180°]$. Moreover, in contrast to previous event-based vision learning works which use small datasets, we show results on the largest and most challenging (due to scene variability) event-based dataset to date.

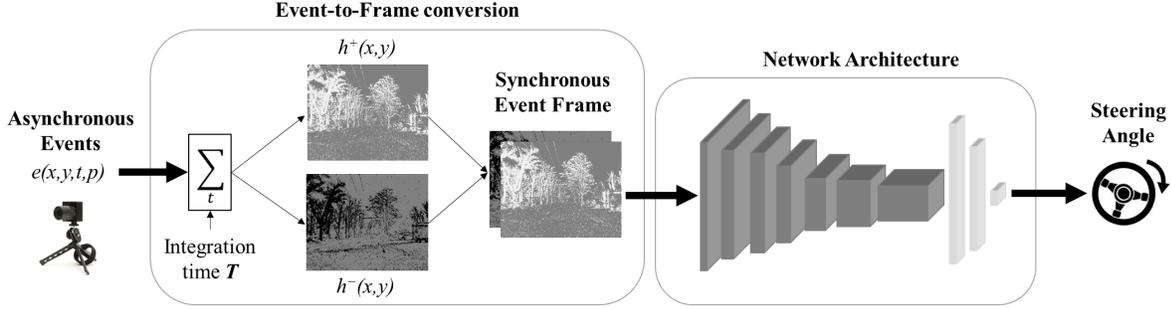

Figure 2: Block diagram of the proposed approach. The output of the event camera is collected into frames over a specified time interval $T$, using a separate channel depending on the event polarity (positive and negative). The resulting synchronous event frames are processed by a ResNet-inspired network, which produces a prediction of the steering angle of the vehicle.

## 3. Methodology

Our approach aims at predicting steering wheel commands from a forward-looking DVS sensor [1] mounted on a car. As shown in Fig. 2, we propose a learning approach that takes as input the visual information acquired by an event camera and outputs the vehicle's steering angle. The events are converted into *event frames* by pixel-wise accumulation over a constant time interval. Then, a deep neural network maps the event frames to steering angles by solving a regression task. In the following, we detail the different steps of the learning process.

### 3.1. Event-to-Frame Conversion

All recent and successful deep learning algorithms are designed for traditional video input data (i.e., frame-based and synchronous) to benefit from conventional processors. In order to take advantage of such techniques, asynchronous events need to be converted into synchronous frames. To do that, we accumulate the events[1] $e_k = (x_k, y_k, t_k, p_k)$ over a given time interval $T$ in a pixel-wise manner, obtaining 2D histograms of events. Since event cameras naturally respond to moving edges, these histograms of events are maps encoding the relative motion between the event camera and the scene. Additionally, due to the sensing principle of event cameras, they are free from redundancy.

Inspired by [18], we use separate histograms for positive and negative events. The histogram for positive events is

$$h^+(x,y) \doteq \sum_{t_k \in T, p_k = +1} \delta(x - x_k, y - y_k), \quad (1)$$

where $\delta$ is the Kronecker delta, and the histogram $h^-$ for the negative events is defined similarly, using $p_k = -1$. The histograms $h^+$ and $h^-$ are stacked to produce a two-channel event image. Events of different polarity are stored in different channels, as opposed to a single channel with the balance of polarities $(h^+ - h^-)$, to avoid information loss due to cancellation in case events of opposite polarity occur in the same pixel during the integration interval $T$.

### 3.2. Learning Approach

**3.2.1. Preprocessing.** A correct normalization of input and output data is essential for reliably training any neural network. Since roads are almost always straight, the steering angle's distribution of a driving car is mainly picked in $[-5°, 5°]$. This unbalanced distribution results in a biased regression. In addition, vehicles frequently stand still because they are exposed, for example, to traffic lights and pedestrians. In those situations where there is no motion, only noisy events will be produced. To handle those problems, we pre-processed the output variable (*i.e.* steering angles) to allow successful learning. To cope with the first issue, only 30 % of the data corresponding to a steering angle lower than 5° is deployed at training time. For the latter we filtered out data corresponding to a vehicle's speed smaller than 20 km h$^{-1}$. To remove outliers, the filtered steering angles are then trimmed at three times their standard deviation and normalized to the range $[-1, 1]$. At testing time, all data corresponding to a steering angle lower than 5° is considered, as well as scenarios under 20 km h$^{-1}$. The regressed steering angles are denormalized to output values in the range $[-180°, 180°]$. Finally, we scaled the network input (i.e., event images) to the range $[0, 1]$.

**3.2.2. Network Architecture.** To unlock the power of convolutional architectures for our study case, we first have to adapt them to accommodate the output of the event camera. Initially, we stack event frames of different polarity, creating a 2D *event image*. Afterwards, we deploy a series of ResNet architectures, i.e., ResNet18 and ResNet50, since they have proved to be easier to optimize as the number of layers (depth) increases, and to better cope with overfitting [11]. As these networks have been designed for image classification purposes, we use them as feature extractors for our regression problem, considering only their convolutional layers. To encode the image features extracted from the last convolutional layer into a vectorized descriptor, we use a global average pooling layer [23] that returns the features' channel-wise mean. This choice has proved to improve the performance, compared to directly adding a fully-connected layer, since it minimizes overfitting by reducing the total number of parameters and it better propagates the gradients. After the global average pooling, we add a fully-connected (FC) layer (256-dimensional for ResNet18 and 1024-dimensional for ResNet50), followed by a ReLU non-linearity and the final one-dimensional fully-connected layer to output the predicted steering angle.

---
[1] An event $e_k$ consists of the spatiotemporal coordinates $(x_k, y_k, t_k)$ of a relative brightness change of predefined magnitude together with its polarity $p_k \in \{-1, +1\}$ (i.e., the sign of the brightness change).

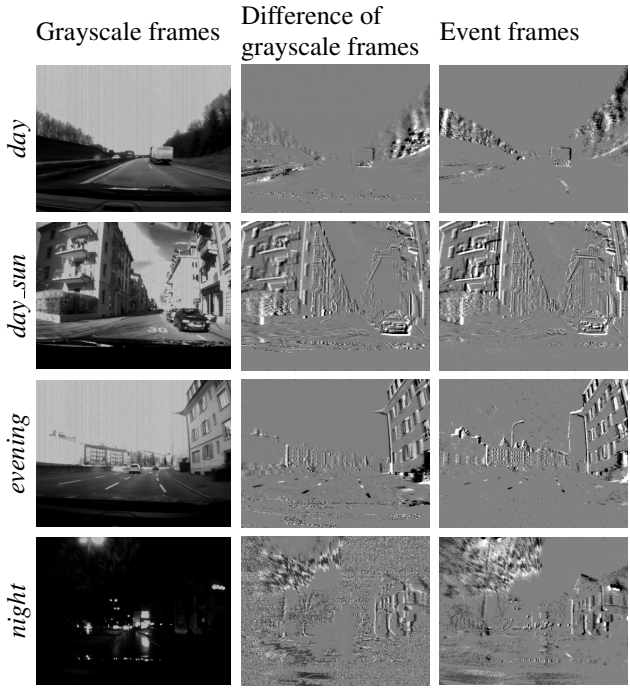

Figure 3: Different input data extracted from the DDD17 dataset [22] for the four lighting conditions. The first column depicts grayscale images collected with a traditional camera (frames of the DAVIS). The second column shows images created by difference of two consecutive grayscale images. The third column corresponds to time-integrated event images (displayed as single-channel images, $h^+ - h^-$, to highlight the similarity with those on the second column).

## 4. Experimental Setup

### 4.1. Dataset

To predict steering angles from event images we use the publicly available DAVIS Driving Dataset 2017 (DDD17) [22]. It contains approximately 12 hours of annotated driving recordings (for a total of 432 GB) collected by a car under different and challenging weather, road and illumination conditions. The dataset includes asynchronous events as well as synchronous, grayscale frames, collected concurrently by the DAVIS[2] sensor [20]. We divided the recordings into four subsets, according to the labels provided by the dataset's authors: *day*, *day_sun*, *evening*, and *night*. Subsets differ not only in the illumination and weather conditions, but also in the route travelled. Fig. 3 depicts some data samples.

Since, while driving, subsequent frames are usually very similar and have almost identical steering angles, randomly dividing the dataset into training and test subsets would result in over-optimistic estimates. Therefore, to properly test generalization of the learned models, we divide the dataset as follows. We split the recordings into consecutive and non-overlapping short sequences of a few seconds each, and use alternate subsets of these sequences for training and testing. In particular, training sequences correspond to 40 seconds of recording, while test sequences to 20 seconds. As shown in Fig. 4, training subsets alternate with test subsets, resulting in different samples.

### 4.2. Performance Comparison on Different Types of Images

We predicted steering angles using three different types of visual inputs:

1. grayscale images,
2. difference of grayscale images,
3. images created by event accumulation.

The grayscale images correspond to absolute intensity frames from a traditional camera (first column of Fig. 3) and they correspond to the typical input of state-of-the-art steering angle prediction systems (Section 2). As already pointed out above, the grayscale frames coming from a DAVIS sensor allow a fair comparison with events, since, being produced by the same photodiodes, they observe exactly the same scene. We also compared our methodology against temporal difference of intensity images (second column of Fig. 3). As shown in the figure, they are similar to the event images (third column of Fig. 3). Intensity differences incorporate temporal information, and, as it will be shown in the experiments (Section 5), they provide a stronger baseline than the absolute intensity images for comparing the results of the event-data architecture. More specifically, event cameras report pixel-wise log-brightness changes of predefined size $C$:

$$L(t) - L(t - \Delta t) = \pm C, \qquad (2)$$

with $L(t) = \log I(t)$, and $I(t)$ being the intensity on the image plane. When these changes (2) are aggregated over some time interval, they quantify the amount of brightness change (increase or decrease) that happened at each pixel,

$$\Delta L \approx (h^+ - h^-) C. \qquad (3)$$

---

[2] The DAVIS camera consists of a traditional grayscale camera and an event sensor (DVS) on the same pixel array, with $346 \times 260$ pixel resolution (DDD17 dataset). Event data is produced simultaneously from the same photodiodes which give the frame-based intensity read-out.

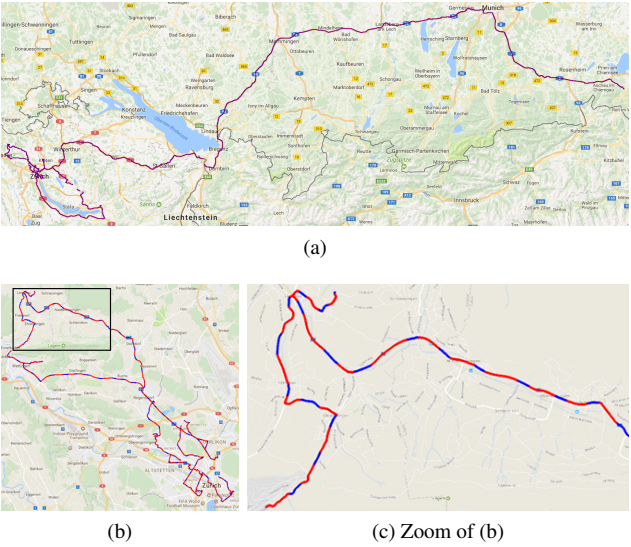

Figure 4: Maps of dataset showing (a) the whole recorded route, covering 1000 km of different roads in Switzerland and Germany; (b) and (c) training and testing frames, in red and blue respectively, for the *day_sun* subset (3 hours of driving). The training sections are different from the testing sections, corresponding to 40 and 20 seconds of recording respectively.

For a small time interval, the difference of two consecutive grayscale frames is a first order (Taylor) approximation to such an intensity change (3) since

$$L(t) - L(t - \Delta t) \doteq \Delta L \approx \frac{\partial L}{\partial t} \Delta t. \qquad (4)$$

This is why images in the second and third columns of Fig. 3, which basically encode the temporal brightness changes over a specified time interval, look similar.

As in [9], we select as ground truth steering wheel angle the one at 1/3 s in the future with respect to the current frame (either event or grayscale frame).

### 4.3. Performance Metrics

Our network addresses the prediction of the steering angle as a regression problem. To evaluate its performance, we use the root-mean-squared error (RMSE) and the explained variance (EVA). The RMSE (5) measures the average magnitude of the prediction error, indicating how close the observed values $\alpha$ are to those predicted by the network $\hat{\alpha}$,

$$\text{RMSE} \doteq \sqrt{\frac{1}{N} \sum_{j=1}^{N} (\hat{\alpha}_j - \alpha_j)^2}. \qquad (5)$$

The EVA (6) measures the proportion of variation in the predicted values with respect to that of the observed values. Such variations are given by the variance of the residuals $\text{Var}(\hat{\alpha} - \alpha)$ and the variance of the observed values $\text{Var}(\alpha)$.

$$\text{EVA} \doteq 1 - \frac{\text{Var}(\hat{\alpha} - \alpha)}{\text{Var}(\alpha)}. \qquad (6)$$

If predicted values approximate the observed values well, the residual variance will be less than the total variance, resulting in $\text{EVA} \lessapprox 1$. Otherwise, the residual variance will be equal or greater than the total variance, producing $\text{EVA} = 0$ or $\text{EVA} < 0$, respectively.

## 5. Experiments

We designed our experiments in order to investigate the following questions:

1. What is the influence of the event integration time, used to produce event frames, on the system performance?
2. What are the advantages of using event images over grayscale or grayscale-difference as input to a network?
3. Does our method scale to very large input spaces? And how does it compare to state-of-the-art methods based on traditional cameras?

To answer the first question, we analyze the performance of our system over a range of integration times (Section 5.1). With regard to the second question, we conduct an extensive study on the four dataset's subsets detailed in Section 4.1 and highlight the advantages of event images over grayscale ones (Section 5.2). Finally, we answer the last questions by learning a single network over the entire dataset (Section 5.3). We show that, despite the large variety of illumination, weather, and road conditions, we can learn a robust and accurate regressor that outperforms state-of-the-art methods based on traditional frames.

### 5.1. Sensitivity Analysis with Respect to the Event Integration Time

In this section, we analyze the performance of the network as a function of the integration time used to generate the input event images from the event stream (Section 3.1). A visual comparison between the input event images for 10, 25, 50, 100, and 200 ms, is shown in Fig. 5. These integration times were chosen to be approximately equispaced in logarithmic scale. It can be observed that the larger the integration time, the larger is the trace of events appearing at the contours of objects. This is due to the fact that they moved a longer distance on the image plane during that time. We hypothesize that the network exploits such motion cues to provide a reliable steering prediction.

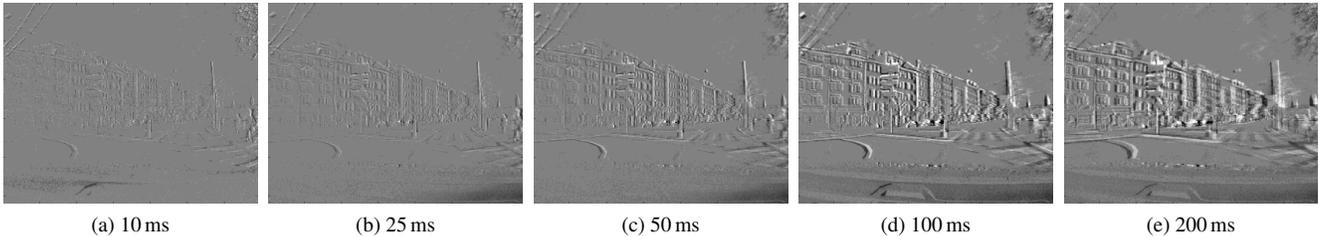

(a) 10 ms    (b) 25 ms    (c) 50 ms    (d) 100 ms    (e) 200 ms

Figure 5: Events collected for different durations of the interval $T$ (cf. Fig. 2). The scene corresponds to a *day_sun* sequence, with the car turning left in an urban environment.

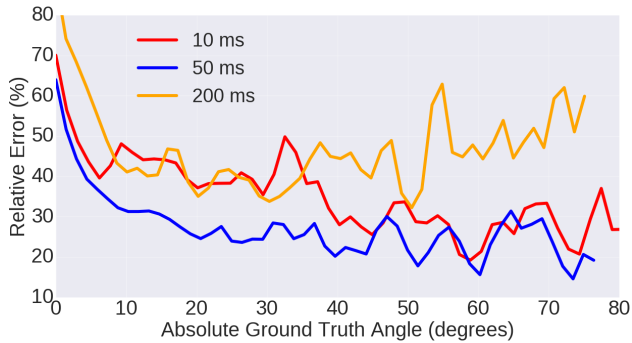

Figure 6: Distribution of the relative error in steering angle prediction as a function of the ground truth steering angle and the event integration time. The performance of the network trained on large integration time (200 ms) degrades for large steering angles. In contrast, the network trained on small integration time (10 ms) predicts well large steering angles, but its performance degrades for smaller angles. The network with an intermediate integration time (50 ms) performs best for large and moderate steering angles. For small angles (<5°), small absolute errors in the angle produce large relative errors regardless of the integration time.

| Integration time $T$ | EVA | RMSE |
|---|---|---|
| 10 ms | 0.790 | 11.53° |
| 25 ms | 0.792 | 10.42° |
| **50 ms** | **0.805** | **9.47°** |
| 100 ms | 0.634 | 13.43° |
| 200 ms | 0.457 | 15.87° |

Table 1: Comparison of the ResNet50 performance for the different integration times on *day_sun* subset.

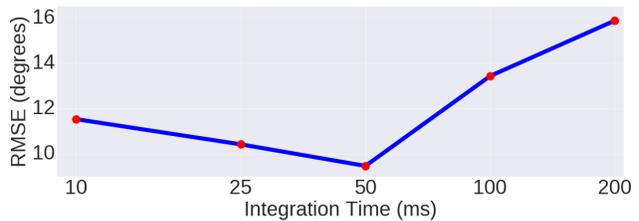

Figure 7: Variation of the RMSE of ResNet50 with respect to the event accumulation time (3rd column of Table 1).

When the integration time is small, event images are generally created out of few events only. Therefore, images associated to relatively small motion are not very discriminative and easy to learn from (Fig. 5a). Conversely, in images created with a long integration time (Fig. 5e) large motion blur washes out the contours of objects, in particular when the car motion is relatively high. Consequently, the produced images lose the discriminability necessary for a reliable estimation. This correlation between the integration time and quality of prediction can be observed in Fig. 6.

Table 1 and Fig. 7 report quantitative results of our ResNet50 network using the five different integration times. As can be observed, the network performs best when it is trained on event images corresponding to 50 ms, and the performance gracefully degrades for smaller and larger integration times. Therefore, in the following experiments we set the integration time to the best value, 50 ms, and further analyze the performance of our method.

### 5.2. Results on Different Illumination Scenarios

Now fixing the integration time to 50 ms, we perform an extensive study to evaluate the advantages of event frames over grayscale-based ones for different parts of the day. To do so, we provide a cross evaluation between architectures and types of input frames in Tables 2 to 5. For fair comparison, we deploy the same convolutional network architectures as feature encoders for all considered inputs, but we train each network independently.

It is interesting to notice that the average RMSE is slightly diverse among different sets. This is to be expected, since RMSE, being dependent on the absolute value of the steering ground truth, is not a good metric for cross comparison between sequences. On the other hand, our second metric, EVA, gives a better way to compare the quality of the learned estimator across different sequences.

The *day* subset, whose results are shown in Table 2, is the most difficult one of the considered partitions. It includes five hours of driving in both urban (including parking lots) and countryside scenarios. The very large variance in the input data (a.k.a. state space) made convergence difficult for the grayscale baseline. In fact, the shallower model, ResNet18, learned on it only a quasi-constant solution (EVA ≈ 0), therefore converging to the data average. In contrast, our method, based on event images, always converged to a solution outperforming the baselines (grayscale and grayscale difference) with both architectures. Interestingly, we observe a very large performance gap between the grayscale difference and the event images for the ResNet18 architecture. The main reasons behind this behavior that we identified are: (*i*) abrupt changes in lighting conditions occasionally produced artifacts in grayscale images (and therefore also in their differences), and (*ii*) at high velocities, grayscale images get blurred and their difference becomes also very noisy (see, e.g., the first column in Fig. 3). Note, however, that the ResNet50 architecture produced a significant performance improvement for both baselines (grayscale images and difference of grayscale images). This is to be expected, since deeper architectures have more training parameters, and can therefore cope better with larger and more complicated state spaces.

A very similar pattern can be observed in the other considered scenarios. Contrary to what we expected, we did not notice a very large degradation of the baselines' performance when considering more challenging illumination conditions as in the *evening* and *night* sequences. However, those latter subsets are much smaller than the other two. Therefore, given the smaller state spaces, the networks have an easier job to model the statistics of the datasets.

As it can be observed in Tables 2 to 5, the event camera solution largely outperforms the baselines on all the analyzed scenarios (best results per row are highlighted in bold). In fact, our proposed methodology consistently achieves very competitive results, even with the simpler ResNet18 architecture.

### 5.3. Results on the Entire Dataset

To evaluate the ability of our proposed methodology to cope with large variations in illumination, driving and weather conditions, we trained a single regressor over the entire dataset. We compare our approach to state-of-the-art architectures that use traditional frames as input: (*i*) Bojarski et al. [15] and (*ii*) the CNN-LSTM architecture, advocated in Xu et al. [9], but without the additional segmentation loss that is not available in our dataset. In our evaluation we do not consider [8], since, in spite of offering an interpretable solution, it gives almost no improvements over the simpler architecture in [15].

Table 6 summarizes the findings of our experiments. In terms of EVA and RMSE, the first baseline obtains a poor performance on the regression task. Indeed, the EVA is very small (0.161) and the RMSE is very close to the dominant steering deviation in the dataset ($\pm 10°$). To provide a stronger baseline, we incorporated temporal information to the grayscale frames by using a CNN-LSTM architecture (resulting EVA $\approx 0.3$). We chose this architecture because it has been reported to provide very competitive results in the evaluation of [9]. This is a more fair comparison because event images inherently contain temporal information.

| Architecture | Grayscale EVA | Grayscale RMSE | Grayscale diff. EVA | Grayscale diff. RMSE | Events EVA | Events RMSE |
|---|---|---|---|---|---|---|
| ResNet18 | 0.047 | 4.57° | 0.329 | 3.65° | **0.551** | **2.99°** |
| ResNet50 | 0.449 | 3.31° | 0.653 | 2.62° | **0.728** | **2.33°** |

Table 2: **Results for *day* subset.**

| Architecture | Grayscale EVA | Grayscale RMSE | Grayscale diff. EVA | Grayscale diff. RMSE | Events EVA | Events RMSE |
|---|---|---|---|---|---|---|
| ResNet18 | 0.125 | 20.07° | 0.729 | 11.53° | **0.742** | **10.87°** |
| ResNet50 | 0.383 | 16.85° | 0.802 | 9.62° | **0.805** | **9.47°** |

Table 3: **Results for *day_sun* subset.**

| Architecture | Grayscale EVA | Grayscale RMSE | Grayscale diff. EVA | Grayscale diff. RMSE | Events EVA | Events RMSE |
|---|---|---|---|---|---|---|
| ResNet18 | 0.172 | 7.23° | 0.183 | 7.19° | **0.518** | **5.45°** |
| ResNet50 | 0.360 | 6.37° | 0.418 | 6.07° | **0.602** | **5.01°** |

Table 4: **Results for *evening* subset.**

| Architecture | Grayscale EVA | Grayscale RMSE | Grayscale diff. EVA | Grayscale diff. RMSE | Events EVA | Events RMSE |
|---|---|---|---|---|---|---|
| ResNet18 | 0.181 | 6.96° | 0.449 | 5.73° | **0.654** | **4.51°** |
| ResNet50 | 0.418 | 5.88° | 0.621 | 4.73° | **0.753** | **3.82°** |

Table 5: **Results for *night* subset.**

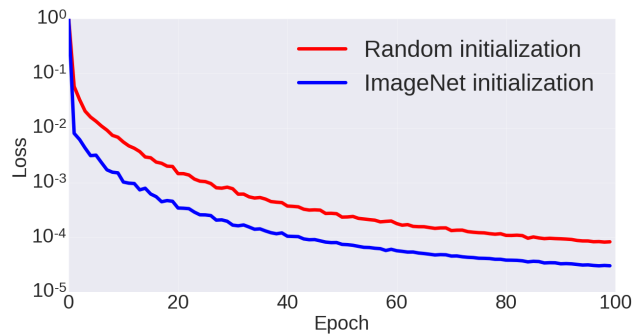

Figure 8: Comparison of training losses for ResNet50 with and without ImageNet initialization.

| Architecture | EVA | RMSE | Input |
|---|---|---|---|
| Bojarski *et al.* [15] | 0.161 | 9.02° | Grayscale |
| CNN-LSTM [9] | 0.300 | 8.19° | Grayscale |
| (Ours) ResNet18 | 0.783 | 4.58° | Events |
| (Ours) ResNet50 (ImageNet init) | **0.826** | **4.10°** | **Events** |
| (Ours) ResNet50 (Random init) | 0.800 | 4.40° | Events |

Table 6: Comparison between two state-of-the-art learning approaches using grayscale frames [9, 15] and the proposed networks that process event frames, for the whole dataset [22]. For ResNet50, both random and ImageNet initializations have been evaluated.

All our proposed architectures based on event images largely outperform the considered baselines based on traditional frames. As it could be reasonably expected, the best results are obtained with the deepest architectures (ResNet50). More interestingly, we noticed some benefits when initializing the ResNet50 weights with those learned on the ImageNet challenge[3] [12]. Even though it is well known that feature learning is generally transferable for different tasks [24], it is still remarkable that parameters learned on traditional RGB images have a positive transfer on time-integrated event images. As pointed out in [24], the main reason behind this is that the first convolutional weights of a network trained on ImageNet are sensitive to low-level features present in the image (e.g., edges), which are present on both traditional and event frames. Leveraging transfer learning from the immense ImageNet classification dataset not only makes training easier and faster (Fig. 8), but also produces a better estimator (Table 6).

## 6. Discussion

A great deal of why a network produces better results on event images than on grayscale frames (or their difference) is their ability to capture scene dynamics. At high velocities, grayscale frames suffer from motion blur (e.g., side-road trees on the first row of Fig. 3), whereas event images preserve edge details due to the very high temporal resolution (microsecond) of event cameras and the fact that we acquire positive and negative events in separate channels that are fed to the network, thus avoiding loss of information (Section 3.1). The temporal aggregation needed to feed the network does, however, affect latency. Additionally, event cameras have a very high dynamic range (HDR) (140 dB compared to the 55 dB range of the grayscale frames in the dataset [22]). Hence, event data represent HDR content of the scene, which is not possible in traditional cameras since that would require long exposure times. This is beneficial in order to be robust to different illumination conditions (bright day, dark night, abrupt transitions in tunnels, etc.). Additionally, since event cameras respond to moving edges and therefore filter out temporally-redundant data, they are more informative about the vehicle motion than individual grayscale frames. As shown qualitatively in Fig 1 and more quantitatively in Fig 6, focusing on moving edges facilitate solving the learning problem. Selecting a good integration time for creating event images additionally improves performance (Fig. 7). Interesting future work concerning this problem is to use reinforcement learning techniques [25] to produce an adaptive integration time policy that depends on the car's speed and the observed scene.

---
[3]To reuse the weights, we averaged the filters of the first convolutional layer along the channel dimension and duplicated them to convert from 3-channel to 2-channel inputs.

State-of-the-art convolutional networks need lots of data to pick up on important motion features. To simplify the task, we showed that it is possible to transfer knowledge from networks trained with traditional images on classification tasks. As a result, we were able to unlock the capabilities of event cameras to solve the task at hand.

## 7. Conclusion

In this work, we showed how a DL-based approach can benefit from the natural response of event cameras to motion and accurately predict a car steering angle under a wide range of conditions. Our DL approach, specifically designed to work with the output of event sensors, learns to predict steering angles by picking them up from the motion cues contained in event-frames. Experimental results showed the robustness of the proposed method, especially under those conditions where grayscale frames fail, e.g., large input spaces, challenging illumination conditions, and fast motion. In conclusion, we showed that it outperforms other state-of-the-art systems based on traditional cameras. We encourage the reader to watch the accompanying video, available at https://youtu.be/_r_bsjkJTHA.

## Acknowledgement

This project was funded by the Swiss National Center of Competence Research (NCCR) Robotics, through the Swiss National Science Foundation, and the SNSF-ERC starting grant. This work has also been partially supported by the Ministerio de Economía, Industria y Competitividad (AEI/FEDER) of the Spanish Government under project TEC2016-75981 (IVME). The Titan Xp used for this research was donated by the NVIDIA Corporation.